\documentclass{article}

\usepackage{microtype}
\usepackage{graphicx}
\usepackage{subcaption}
\usepackage{booktabs} 

\usepackage{hyperref}

\usepackage[accepted]{icml2025}

\usepackage{amsmath}
\usepackage{amssymb}
\usepackage{mathtools}
\usepackage{amsthm}
\usepackage{graphicx}
\usepackage{subcaption}
\usepackage{booktabs}
\usepackage{multirow}
\usepackage{enumitem}
\usepackage{float}
\usepackage{comment}

\newcommand{\bp}{\mathcal{B}} 
\newcommand{\task}{\mathcal{T}} 
\newcommand{\slm}{\mathcal{M}_{slm}} 
\newcommand{\llm}{\mathcal{M}_{llm}} 

\newcommand{\phimini}{Phi3-mini}
\newcommand{\gptmini}{GPT4o-mini}
\newcommand{\gpt}{GPT4o}
\newcommand{\mistral}{Mistral-7B}

\newcommand{\concreteexample}{\texttt{concrete-example}}
\newcommand{\abstractexample}{\texttt{abstract-example}}
\newcommand{\detailed}{\texttt{detailed-pattern}}
\newcommand{\plain}{\texttt{plain-pattern}}
\newcommand{\concise}{\texttt{concise-highlevel}}
\newcommand{\instruction}{\texttt{instruction-focus}}
\newcommand{\contextual}{\texttt{contextual-explanation}}
\newcommand{\reflective}{\texttt{reflective-refinement}}

\newcommand{\bulletpoint}{\texttt{bullet-points}}
\newcommand{\decision}{\texttt{decision-making}}
\newcommand{\plan}{\texttt{plan-and-solve}}

\newcommand{\cotoneshot}{CoT (1-shot)}
\newcommand{\cotthreeshot}{CoT (3-shot)}
\newcommand{\apo}{APO}
\newcommand{\bpwoapo}{BP (w.o. APO)}
\newcommand{\bpwapo}{BP (w. APO)}
\newcommand{\bpwapotemplatesearch}{BP (w. APO) + Template Search}
\usepackage{graphicx}

\usepackage[capitalize,noabbrev]{cleveref}

\theoremstyle{plain}

\theoremstyle{definition}

\theoremstyle{remark}

\usepackage[textsize=tiny]{todonotes}

\icmltitlerunning{Enhancing Reasoning Capabilities of Small Language Models with Blueprints and Prompt Template Search}

\begin{document}

\twocolumn[
\icmltitle{Enhancing Reasoning Capabilities of Small Language Models with \\Blueprints and Prompt Template Search}




\begin{icmlauthorlist}
\icmlauthor{Dongge Han}{comp}
\icmlauthor{Menglin Xia}{comp}
\icmlauthor{Daniel Madrigal Diaz}{comp}
\icmlauthor{Samuel Kessler}{comp}
\icmlauthor{Ankur Mallick}{comp}
\icmlauthor{Xuchao Zhang}{comp}
\icmlauthor{Mirian Del Carmen Hipolito Garcia}{comp}
\icmlauthor{Jin Xu}{comp}
\icmlauthor{Victor Rühle}{comp}
\icmlauthor{Saravan Rajmohan}{comp}
\end{icmlauthorlist}

\icmlaffiliation{comp}{Microsoft}
\icmlcorrespondingauthor{Dongge Han}{\href{mailto:donggehan@microsoft.com}{donggehan@microsoft.com}}

\icmlkeywords{Small Language Model, LLM, Efficiency, Automatic Prompt Optimization}

\vskip 0.3in
]



\printAffiliationsAndNotice{}  

\begin{abstract}
Small language models (SLMs) offer promising and efficient alternatives to large language models (LLMs). However, SLMs' limited capacity restricts their reasoning capabilities and makes them sensitive to prompt variations. To address these challenges, we propose a novel framework that enhances SLM reasoning capabilities through LLM generated blueprints. The blueprints provide structured, high-level reasoning guides that help SLMs systematically tackle related problems. Furthermore, our framework integrates a prompt template search mechanism to mitigate the SLMs' sensitivity to prompt variations. Our framework demonstrates improved SLM performance across various tasks, including math (GSM8K), coding (MBPP), and logic reasoning (BBH). Our approach improves the reasoning capabilities of SLMs without increasing model size or requiring additional training, offering a lightweight and deployment-friendly solution for on-device or resource-constrained environments.
\end{abstract}

\begin{figure*}[t]
    \centering
     \captionsetup{skip=2pt}
    \includegraphics[width=\linewidth]{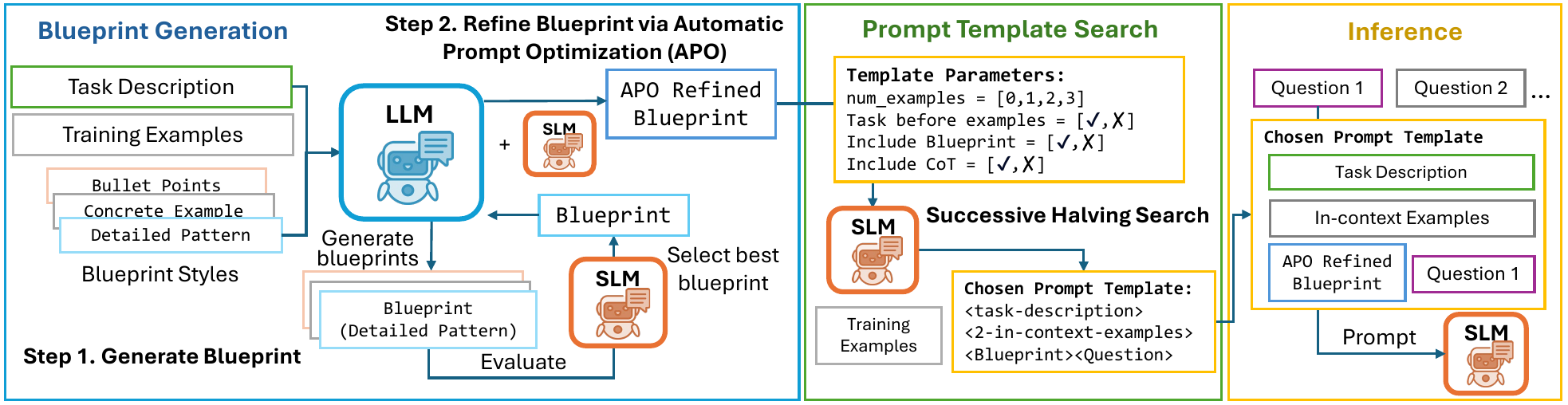}
    \caption{Illustration of our framework. During training, for each SLM and task category, an LLM is used to generate blueprints using various input styles, from which the best blueprint is picked, and refined via APO. Then we perform template search to find the best prompt template. During inference, the chosen blueprint and template are reused across problems within the task category to form the input prompt to the SLM.}
    \label{fig:framework}
\end{figure*}

\section{Introduction}
Large language models (LLMs) have demonstrated remarkable capabilities across a wide range of real-world applications. However, deploying LLMs often requires substantial computational resources, posing challenges in efficiency and their general applicability~\cite{wan2023efficient,chowdhery2023palm,du2022glam}. Small language models (SLMs)~\cite{gpt4omini, jiang2023mistral} such as \phimini{} with only 3.8B parameters~\cite{abdin2024phi}, offer promising alternatives by providing efficient solutions that are also more accessible in resource-constrained environments.

Despite these advantages, SLMs still face challenges that hinder their broader adoption. Due to their limited capacity, SLMs exhibit less reasoning capabilities such as in chain-of-thought (CoT) and in-context learning, as reported by~\cite{magister2022teaching, wei2023larger,brown2020language}. 
For example, unlike LLMs, which can effectively extract abstract chain-of-thought reasoning patterns from the in-context examples and apply to new problems, SLMs’ limited reasoning capacity sometimes make it challenging for them to effectively extract and transfer insights from the examples to new problems, limiting their generalization capabilities.

To address these challenges, we propose a novel framework that enhances SLM problem-solving via \emph{blueprints}. A blueprint is a reusable, step-by-step reasoning guide designed to help SLMs solve a class of related problems. The blueprint is generated with an LLM by extracting high-level, abstract reasoning instructions from concrete examples, facilitating SLMs with a clear and actionable plan for tackling complex problems. Unlike CoT or in-context learning, where SLMs may struggle to come up with, or extract from the examples correct reasoning steps and generalize to new problems, blueprints provide explicit reasoning guidance to bridge this gap.  By outlining high-level reasoning steps and sometimes incorporating illustrative examples, blueprints facilitate SLMs with the abstract reasoning steps and how to apply them to solve new problems. Moreover, we tailor the blueprint to each problem class and SLM by selecting from diverse blueprint styles and refining it using Automatic Prompt Optimization (APO)~\cite{pryzant2023automatic}.
In addition to blueprint generation, our framework incorporates a prompt template search mechanism as a further contribution to enhance the consistency of SLM performance, addressing their sensitivity to prompt variations. 

We evaluate our approach on various SLMs - \gptmini{}~\cite{gpt4omini}, \mistral{}~\cite{jiang2023mistral} and \phimini{}, across a diverse set of tasks, including math (GSM8K~\citet{cobbe2021training}), coding (MBPP~\citet{austin2021program}), and logic reasoning (BBH~\citet{suzgun2022challenging}). The experimental results demonstrate that our framework consistently improves SLM performance, outperforming state-of-the-art methods. These findings highlight the effectiveness of blueprint-guided reasoning and prompt template optimization in overcoming the inherent limitations in the reasoning capabilities and sensitivity of SLMs.

\begin{figure}[t]
 \captionsetup{skip=1pt}
    \centering
    \includegraphics[width=\linewidth]{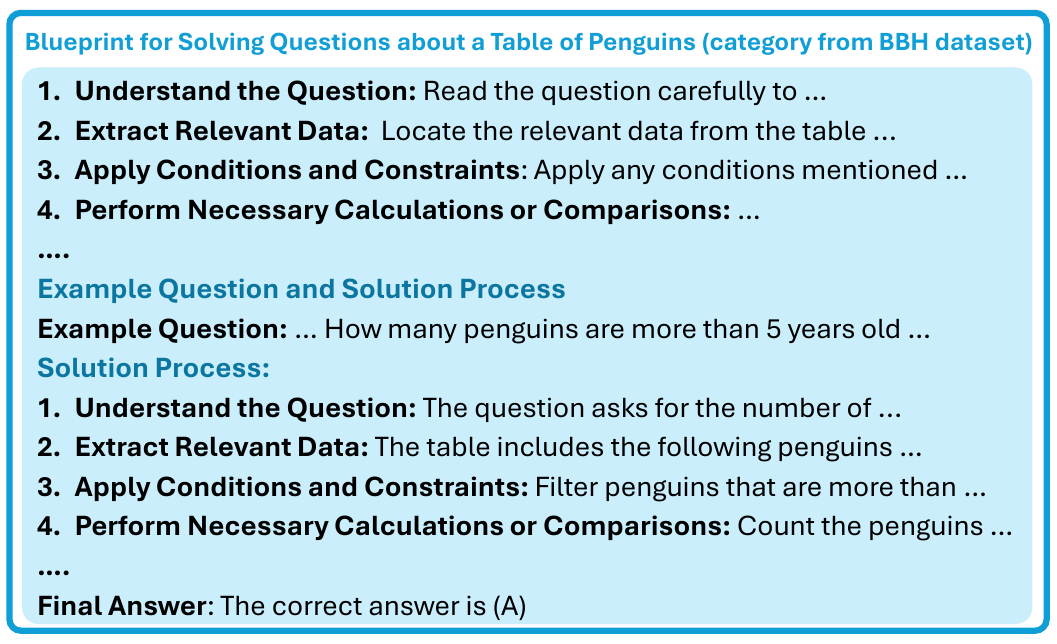}
    \caption{Example LLM generated blueprint. The blueprint includes high-level reasoning steps, and optionally an in-context example question and solution process as a guide for how to apply the reasoning steps.}
    \label{fig:example_blueprint}
\end{figure}
\label{sec:intro}

\section{Related Work}
Prompt engineering is an efficient way to enhance LLMs, including methods like CoT~\cite{wei2022chain}, ReAct~\cite{yao2022react}, ToT~\cite{long2023large}, and few-shot prompting~\cite{brown2020language}, as well as automatic prompt engineering methods~\cite{lester2021power, deng2022rlprompt, ma2024large, yang2024zhongjing} such as APO~\cite{pryzant2023automatic}, APE~\cite{zhou2022large}, PromptBreeder~\cite{fernando2023promptbreeder}, and PromptWizard~\cite{agarwal2024promptwizard}. These methods primarily focus on refining a short, high-level task description for an LLM. In contrast, our framework specifically targets the unique challenges of SLMs, such as limited reasoning capacity and prompt sensitivity, by generating reusable and structured blueprints from example problems and prompt template optimization. These blueprints provide explicit multi-step guidance tailored to SLM needs, offering detailed clarification and actionable reasoning support. While we used APO for blueprint refinement in our experiments, our framework is also versatile to integrate other aforementioned prompt refinement methods. 
\label{sec:related}

\section{Method}
In this section, we present our framework designed to enhance the reasoning capability of SLMs with blueprints and prompt template search.

\begin{figure*}[t]
    \captionsetup{skip=-10pt}
    \begin{subfigure}[b]{0.22\textwidth}
        \includegraphics[width=\textwidth, height=0.8\textwidth]{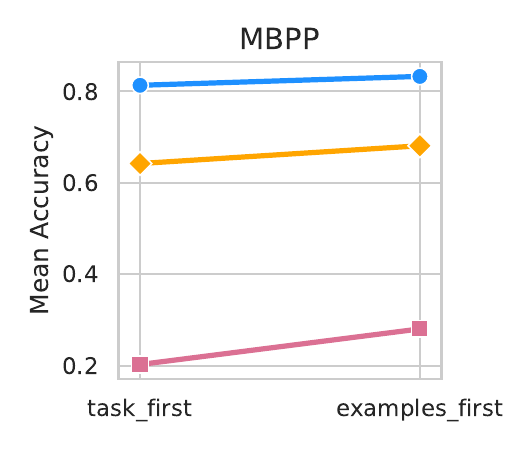}
        \label{fig:order_mbpp}
    \end{subfigure}
    \hspace{-2mm}
    \begin{subfigure}[b]{0.22\textwidth}
        \includegraphics[width=\textwidth, height=0.8\textwidth]{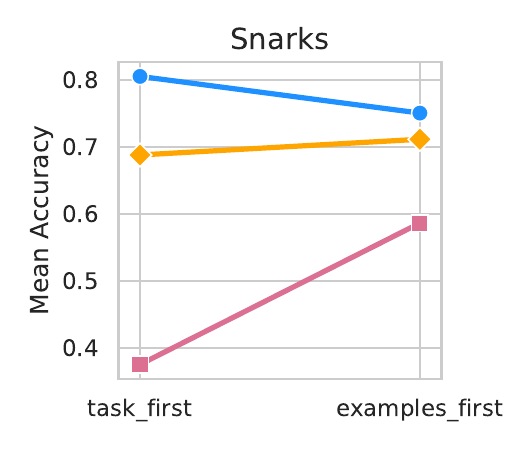}
        \label{fig:order_snarks}
    \end{subfigure}
    \hspace{-2mm}
    \begin{subfigure}[b]{0.22\textwidth}
        \includegraphics[width=\textwidth, height=0.8\textwidth]{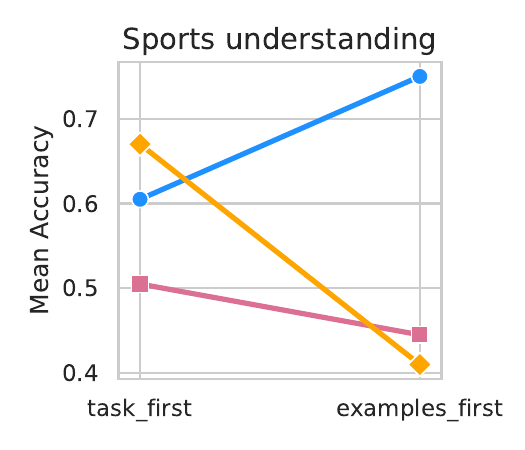}
        \label{fig:order_sports}
    \end{subfigure}
    \hspace{-2mm}
    \begin{subfigure}[b]{0.22\textwidth}
        \includegraphics[width=\textwidth, height=0.8\textwidth]{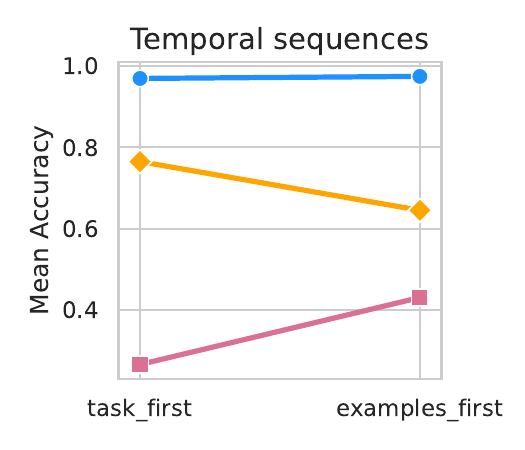}
        \label{fig:order_temporal}
    \end{subfigure}
    \hspace{-2mm}
    \begin{subfigure}[b]{0.1\textwidth}
        \includegraphics[width=\textwidth, height=1.7\textwidth]{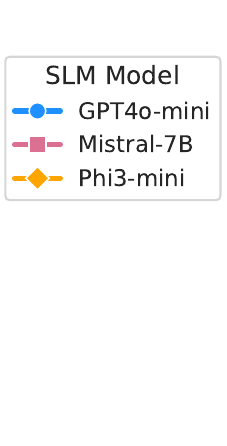}
        \label{fig:order_legend}
    \end{subfigure}
    \caption{Sensitivity of SLMs to prompt variations across tasks - Even simple changes in the template, like swapping the order of \texttt{<task-description>} and \texttt{<in-context-example>}, can significantly affect performance, with impacts varying per SLM and task, e.g., in Sports understanding (Snarks, Sports understanding and Temporal sequences are sub-categories from the BBH dataset), \gptmini{} performed better with examples before the task description, while \mistral{} and \phimini{} preferred the opposite. There is no universal best prompt template for all tasks and SLMs, making task and model-specific prompt optimization necessary. 
    }
    \label{fig:sensitivity}
\end{figure*}

\subsection{Problem Formulation}
\label{subsec: problem_formulation}
Fig.~\ref{fig:framework} shows an overview of our framework.
We refer to a \textbf{\emph{task}} $\task$ as a collection of related \textbf{\emph{problems}} (e.g., a math question) within the same category (e.g., causal judgement, math reasoning). During training, for each task category $\task$ and SLM model $\slm$ (e.g., \phimini{}), we use a small set of examples to generate a \textbf{\emph{blueprint}} that is tailored to guide the SLM with problem-solving for problems within $\task$; we also generate a \textbf{\emph{prompt template}} (e.g., \texttt{<task-description><2-in-context-examples> <blueprint><question>}, which defines the prompt components and their ordering). During inference, the blueprint and prompt template can be efficiently reused to form prompts to $\slm$ across incoming problems within task $\task$.

\subsection{Blueprint Generation and Optimization}
\label{subsec:blueprint}
To address the limited reasoning capabilities of SLMs, we leverage an LLM $\llm$ (we use~\gpt{}~\cite{gpt4o} in this work) to generate a reusable blueprint tailored per SLM per task. The blueprint serves as a structured guide, outlining the general pattern of steps the SLM should follow to arrive at the correct solution, which can be reused across problems within the task category. 

As shown in Fig.~\ref{fig:framework}, we first generate the blueprint, then refine it via automatic prompt optimization (APO). An example generated blueprint is shown in Appendix Fig.~\ref{fig:example_blueprint}.
To generate the blueprint, we begin by sampling $M$ example problems from $D^{\task}_{train}$, each consisting of a question and its step-by-step solution. These examples are concatenated into a single prompt to $\llm$ to generate the initial blueprint that encapsulates the shared reasoning pattern across the example problems. Since different SLMs may prefer different prompt \emph{styles}~\cite{fernando2023promptbreeder, agarwal2024promptwizard} depending on the task, we personalize the blueprint for each SLM by preparing $K$ distinct blueprint generation styles $S = \{S_1, ... S_K\}$. These styles, such as \concise{}, \bulletpoint{}, vary in aspects such as the level of detail and formatting. 
To identify $\slm$'s preferred blueprint style on task $\task$, we augment the blueprint to the $\slm$ prompt and compare its performance on sampled training examples. The blueprint which yields the highest performance is selected.
More details about the styles and blueprint generation prompts are included in Appendix~\ref{sec:appendix_blueprint}.

To further refine the blueprint according to $\slm{}$'s behavior on task $\task$, we employ Automatic Prompt Optimization (APO)~\cite{pryzant2023automatic}. APO iteratively improves the blueprint by evaluating $\slm$ on training examples and identifying errors, which are compiled into an error message. The LLM generates error analyses from this message to update and paraphrase the blueprint, which are then re-evaluated, with the best-performing version selected for the next round of refinement. More details on the APO refinement procedure and prompts used are included in Appendix~\ref{sec:appendix_blueprint_apo}.

\subsection{Prompt Template Search}
\label{subsec:template_search}
SLMs are often sensitive to the prompt format, such as the template and style variations. As shown in Fig.~\ref{fig:sensitivity}, we observed that simply interchanging the task description and in-context examples have a large effect on SLM performance.  
Furthermore, the effects of prompt format changes vary across different SLMs and tasks, indicating that there's not a universally optimal prompt template for all SLMs and tasks. 

Motivated by this sensitivity of SLMs to prompt templates and the variability across different SLMs and tasks, we perform prompt template search to systematically identify the optimal prompt template for each SLM and task, as shown in Fig.~\ref{fig:framework}. To begin with, we define a search space encompassing key prompt template parameters such as whether or not to include a blueprint. To efficiently search the space, We adopt a simplified successive halving search~\cite{pryzant2023automatic}: we begin with evaluating all $N$ possible templates on $k$ sampled training examples, then iteratively narrow down the search space with a reduction factor $f$ (e.g., $f=2$), i.e., retain the $\lfloor\frac{N}{f}\rfloor$ top scoring candidates and re-evaluate, until one prompt template remains. By progressively narrowing down the template candidates, successive halving efficiently identifies the top prompt template for each SLM and task.
\label{sec:method}

\section{Experiments}

\begin{table*}[t]
\footnotesize
\centering
\setlength\tabcolsep{0.5pt}
\begin{tabular}{@{}llccccccccc@{}}
\toprule
 &  & \multicolumn{3}{c}{\textbf{BBH}} & \multicolumn{3}{c}{\textbf{GSM8K}} & \multicolumn{3}{c}{\textbf{MBPP}} \\ \midrule
 &  & \textbf{\gptmini{}} & \textbf{Mistral7B} & \textbf{\phimini{}} & \textbf{\gptmini{}} & \textbf{Mistral7B} & \textbf{\phimini{}} & \textbf{\gptmini{}} & \textbf{Mistral7B} & \textbf{\phimini{}} \\ \midrule
\multirow{3}{*}{\textbf{Baseline} \quad} & \cotoneshot{} & 0.839 & 0.449 & 0.678 & 0.940 & 0.400 & 0.807 & 0.813 & 0.202 & 0.642 \\ 
 & \cotthreeshot{} & 0.850 & 0.53 & 0.686 & 0.923 & 0.347 & 0.780 & 0.821 & 0.233 & 0.658 \\ 
  & \apo{} & 0.825 & 0.435 & 0.651 & 0.930 & 0.407 & 0.797 & \textbf{0.837} & 0.288 & 0.658 \\ 
  \midrule
\multicolumn{1}{c}{\multirow{3}{*}{\textbf{Ours} \quad}} & \bpwoapo{} & 0.882 & 0.539 & \textbf{0.743} & 0.947 & 0.450 & 0.817 & 0.821 & \textbf{0.440} & 0.689 \\ 
\multicolumn{1}{c}{} & \bpwapo{} & \textbf{0.884} & 0.559 & 0.738 & 0.943 & 0.480 & \textbf{0.840} & 0.829 & 0.401 & 0.681 \\ 
\multicolumn{1}{c}{} & BP (w. APO) + TS & \textbf{0.884} & \textbf{0.572} & 0.722 & \textbf{0.953} & \textbf{0.490} & 0.820 & 0.833 & 0.424 & \textbf{0.696} \\ 
\bottomrule
\end{tabular}
\caption{Performance comparison of different methods (rows) across various SLMs and datasets (GSM8K, MBPP, BBH). The table highlights the accuracy improvements of blueprint-based methods (BP) over CoT and APO (with task description) baselines. BP (w. APO) and BP (w.o. APO) refer to blueprint generation with/without further APO refinement, respectively. Both methods consistently outperform the CoT and APO baselines. The BP (w. APO) + TS (Template Search) variant demonstrates the best overall performance, showing optimal task accuracy in 5 out of 9 SLM/dataset combinations and near-optimal performance in the remaining ones.}
\label{table: main}
\end{table*}

\begin{figure*}[h]
    \centering
    \includegraphics[width=\linewidth]{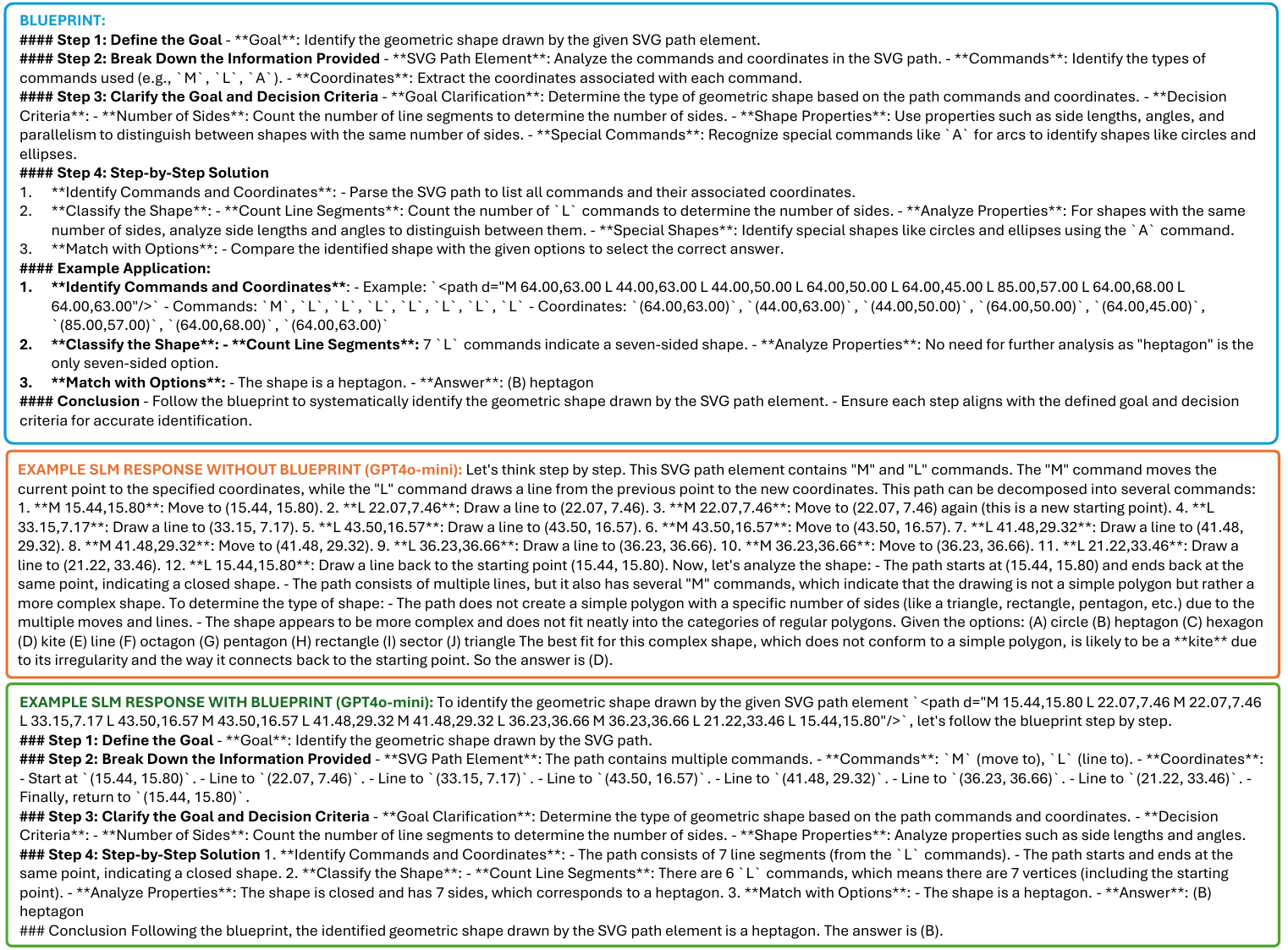}
    \caption{Example blueprint from the BBH geometric shape reasoning task, and example \gptmini{} responses without and with blueprint, respectively. We can observe that the blueprint listed the reasoning steps, as well as showing an example application, which guides the SLM how to apply the blueprint to example problems. The SLM with blueprint indeed followed the reasoning steps as demonstrated in the blueprint. For each high-level step, the SLM first stated the high-level step provided by the blueprint, followed by its application of the abstract step to the concrete problem.}
    \label{fig:blueprint_following}
\end{figure*}

\begin{figure*}[h]
    \begin{subfigure}[b]{0.5\linewidth}
        \includegraphics[width=\linewidth, height=0.6\linewidth]{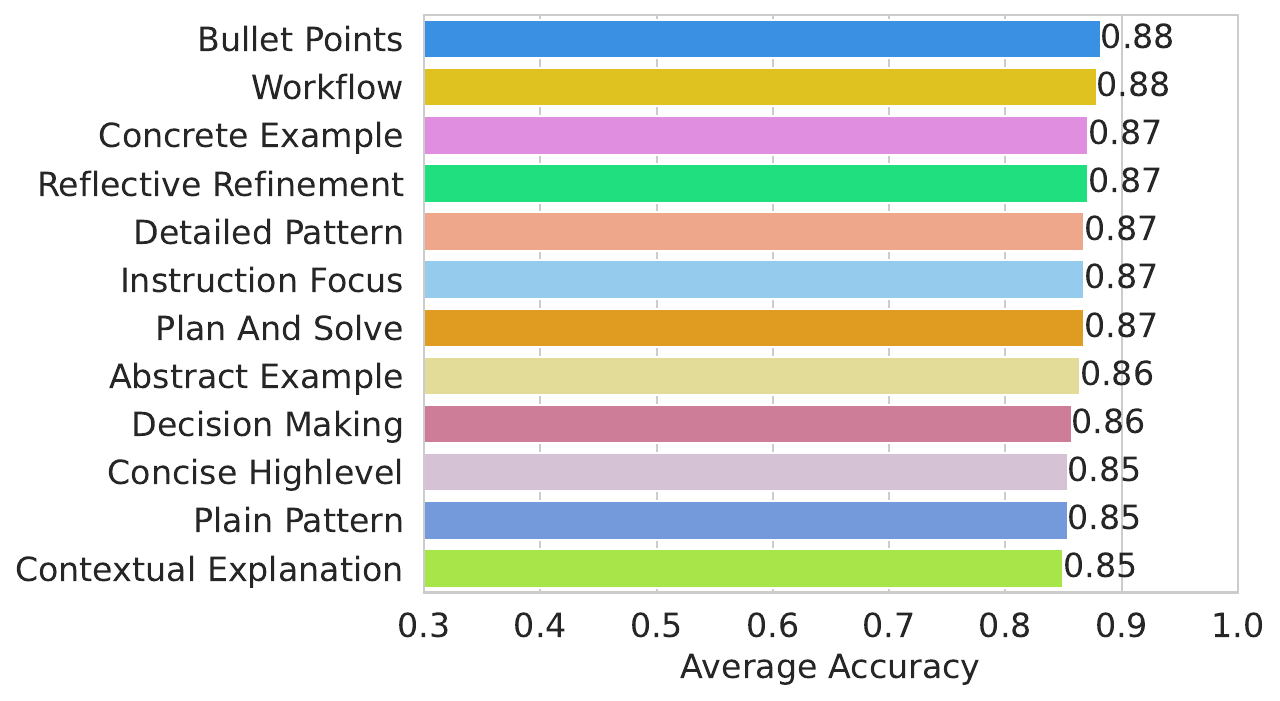}
        \caption{\gptmini{}}
        \label{fig:accuracy_styles_gpt4o_mini}
    \end{subfigure}
    \begin{subfigure}[b]{0.5\linewidth}
        \includegraphics[width=\linewidth, height=0.6\linewidth]{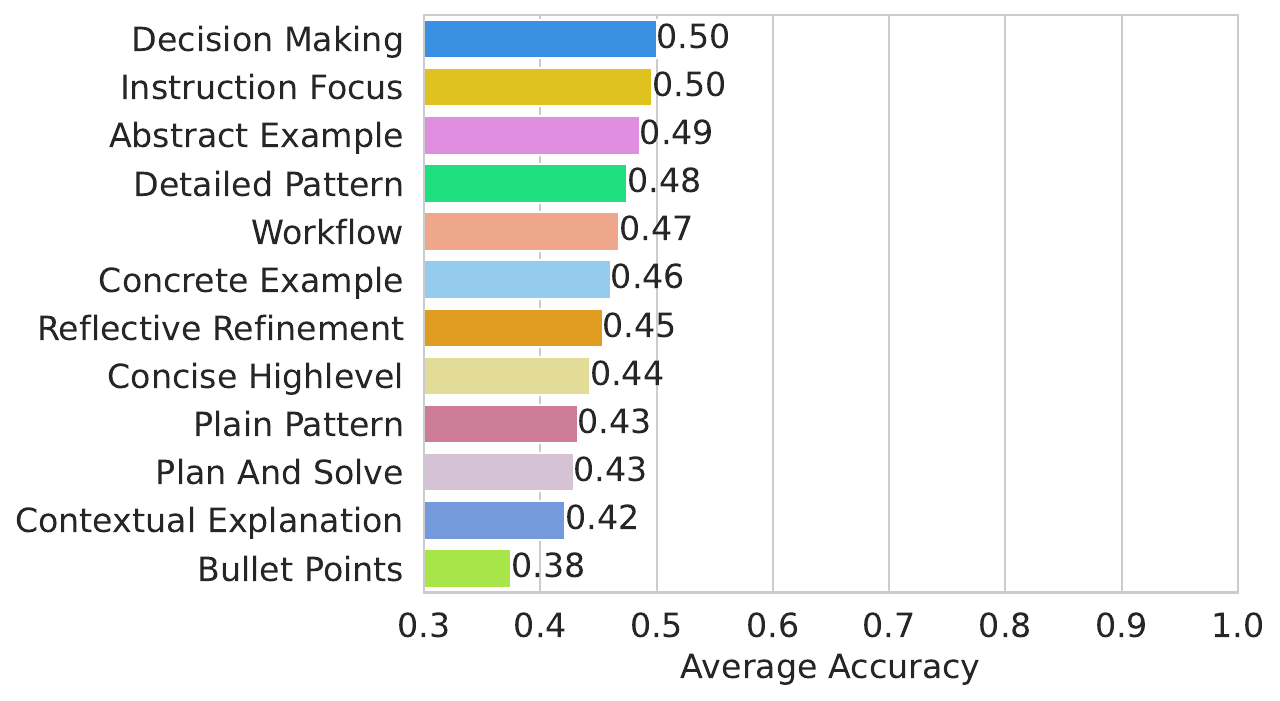}
        \caption{\mistral{}}
        \label{fig:accuracy_styles_mistral_7b}
    \end{subfigure}
    \begin{subfigure}[b]{0.5\linewidth}
        \includegraphics[width=\linewidth, height=0.6\linewidth]{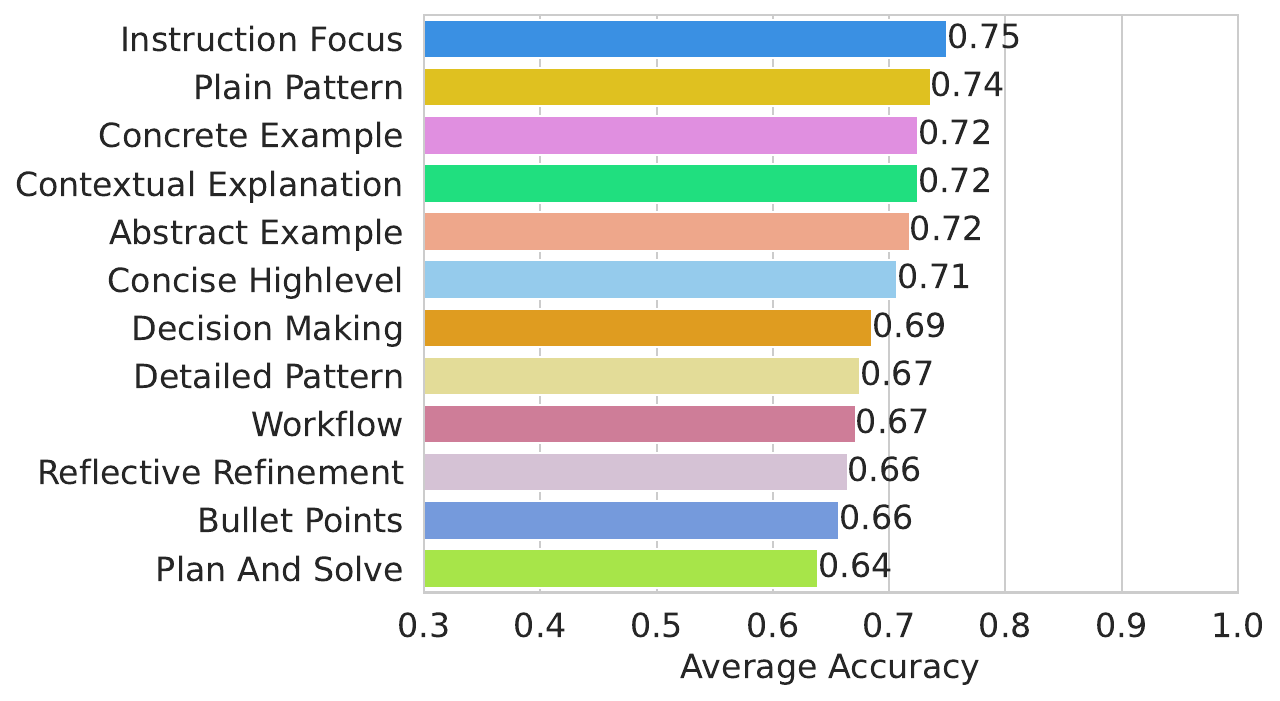}
        \caption{\phimini{}}
        \label{fig:accuracy_styles_phi3_mini}
    \end{subfigure}
    \caption{Accuracy of SLM task performance across blueprint styles. For each blueprint style and SLM, the performance is averaged over 280 examples ($10$ samples per task category for a total of 28 task categories spanning the GSM8K, MBPP and BBH datasets.)}
    \label{fig:accuracy_styles_blueprint}
\end{figure*}

In the experiments, we demonstrate the effectiveness of our proposed framework across various reasoning tasks and SLM models. Followed by the main findings, we present a qualitative example to illustrate the blueprint following behavior, to understand how the blueprint helps guide the SLM. Finally, we present an interesting finding about the different preferences of SLMs to the various blueprint styles.
\subsection{Setup} 
We evaluate the effectiveness of our proposed framework for enhancing SLM reasoning across various reasoning tasks, using three different SLM models \gptmini{}, \phimini{}, and \mistral{} on three widely used different domains: 
\begin{itemize}
    \item GSM8K (Mathematical reasoning)
    \item MBPP (Python coding)
    \item BBH task suite (Logic reasoning, comprising over 20 different task categories)
\end{itemize}
We measured the SLMs' solution accuracy on all datasets. For GSM8K and BBH, we measure the accuracy of the SLM answer compared with ground truth answer. For MBPP, the SLM generated code is correct if it passes all test cases. 
The compared methods are:
1. CoT (1-shot) and 2. CoT (3-shot): chain-of-thought prompting including "Let's think step by step", and 1 (or 3) in-context learning examples; 3. APO: baseline APO method which refines the task description, and 3 variants of our method: 4. BP (w.o. APO): where the SLM uses the top scoring blueprint style without further APO. (BP refers to blueprint); 5. BP (w. APO): using the blueprint refined with APO. Both variants use 1-shot example and no CoT; 6. BP (w. APO) + Template Search, using the APO refined blueprint and prompt template found via template search. For fair comparison, we use the same in-context examples and held-out test set (200 - 300 datapoints for GSM8K, MBPP and each BBH sub-category) across all compared methods. More details on experiment setup can be found in Appendix~\ref{sec:appendix_experiment_details}

\subsection{Main Results}
Table~\ref{table: main} summarizes the performance comparison of the methods for different SLMs across GSM8K, MBPP and BBH. 
In comparison, our method BP (w.o. APO) already outperforms both CoT and APO baselines across almost all datasets and SLMs, e.g., compared with CoT (3-shot), improving accuracy of $+20\%$ for \mistral{} on MBPP, $+5.7\%$ for \phimini{} on BBH, and $+3.2\%$ for \gptmini{} on BBH. With APO, the BP (w. APO) variant slightly improves performance over the non-APO variant in only some categories, e.g., $+2\%$ for \mistral{} on BBH, and $+3\%$ on GSM8K. The BP (w. APO) + Template Search variant shows best performance across all variants, outperforming all other methods in 5 out of 9 SLM/task combinations, and competitive performance in the rest. Note that the template search variant does not always provide the best result due to the small number of examples used to compare the template candidates during template search. These results demonstrate the effectiveness of our proposed framework across a wide range of reasoning tasks and various SLM models.

\subsection{Example Blueprint and SLM Response following Blueprint Steps}
\label{sec:appendix_bp_following}
To understand the format of the generated blueprint and how the SLM follows the blueprint, we show a qualitative example in Fig.~\ref{fig:blueprint_following}. In the figure, we show an example blueprint generated for the logical reasoning tasks for geometric shapes reasoning category (from BBH) for \gptmini{}, and example responses from \gptmini{} before and after following the blueprint, respectively. As shown in the figure, the blueprint starts by listing the reasoning steps, including first defining the goal, then break down the information provided, clarifying the goal and decision criteria, and detailed step-by-step solutions, followed by a conclusion. Each high-level step is illustrated with a description of how to carry out the step, helping the SLM to clearly follow the instruction. After presenting the high-level steps, the blueprint also shows an example application, which guides the SLM on how to apply the blueprint to solve a concrete example problem. 

To see the effectiveness of the blueprint, we show the response from \gptmini{} with (in the green box) and without (in the orange box) the blueprint. Note that without using the blueprint, the SLM answered the problem incorrectly even with chain-of-thought reasoning, while the response from \gptmini{} using the blueprint answered the problem correctly. Comparing the answers, the response without blueprint directly dives into the detailed problem-solving steps, while the response following the blueprint presents better structured reasoning.  We can observe that the SLM indeed followed the reasoning steps as demonstrated in the blueprint. Instead of diving into the detailed problem solution directly, the SLM starts by defining the problem goal, breaks down information gathered, then performs step-by-step solution.

\subsection{SLM performance with different blueprint styles}
\label{sec:appendix_blueprint_style_performance}
To explore the underlying preferences of different SLMs with various blueprint styles, Fig.~\ref{fig:accuracy_styles_blueprint} presents the blueprint styles alongside the average SLM performance when using the blueprints, averaged over all task categories (i.e., 10 examples per blueprint style per category, for a total of 28 categories, including GSM8K, MBPP and all BBH sub-categories). 

Overall, \gptmini{} is demonstrated the highest level of robustness to the blueprint styles, with $3\%$ performance difference between the best and worst performing styles, while \phimini{} and \mistral{} are slightly more sensitive, showing a $11\%$ and $12\%$ performance differences, respectively.

Interestingly, each SLM has different preferences for the blueprint styles -- \gptmini{} slightly prefers instructions in bullet pointed format and illustration with workflows or concrete examples, while it was slightly less effective with \contextual{}, \plain{} and \concise{}. In contrast, \phimini{} preferred \instruction{}, \plain{} and \concreteexample{}, which put stronger focus on clear and actionable guidance, while it was less effective with the multi-step styles such as \reflective{} and \plan{}, and \bulletpoint{} formatted instructions. Finally, \mistral{} preferred blueprint styles with \decision{}, \instruction{} and \abstractexample{}, which provide \mistral{} with clear and actionable guidance, while it was less effective with \plan{}, \contextual{} and \bulletpoint{}. 

The above findings show that the SLMs tend to prefer different styles, e.g., bullet pointed format works better with \gptmini{} while less well with \phimini{} and \mistral{}. Therefore, it is important to match the prompt styles with the preferred format of the SLMs. This observation may extend beyond just blueprints and provide insights to general prompting techniques of SLMs.
For the list of detailed descriptions of all blueprint styles and the corresponding prompts, please refer to Appendix~\ref{sec:appendix_blueprint_style}.


\label{sec:experiments}

\section{Conclusion}
We presented a blueprint generation and prompt template search framework designed to improve the reasoning capabilities of SLMs and addressing their sensitivity to prompt variations. By providing structured guidance, our approach has shown consistent improvements to SLM capabilities across multiple domains (math, coding, reasoning). Our approach effectively improves the reasoning capabilities of SLMs, offering a lightweight solution for on-device or resource-constrained environments. Future works include extending to more domains and SLMs, with the goal of enhancing SLMs on a wide range of applications. 

\section*{Impact Statement}
This work contributes to the growing effort to make generative AI more efficient, accessible, and sustainable by enhancing the reasoning capabilities of SLMs without requiring additional training or model size increases. Our lightweight, training-free framework aligns closely with the mission by enabling more effective use of SLMs in resource-constrained settings. By demonstrating how structured prompting and template selection can significantly boost performance, we help advance the broader goal of achieving competitive, efficient, and privacy-aware alternatives to LLMs. We believe this approach supports responsible, scalable deployment of AI across diverse applications.

\nocite{langley00}

\bibliography{example_paper}
\bibliographystyle{icml2025}

\newpage
\appendix
\onecolumn

\section{Appendix: Blueprint Generation and Optimization}
\label{sec:appendix_blueprint}

\subsection{Blueprint Generation Styles}
\label{sec:appendix_blueprint_style}
\begin{figure*}
    \centering
    \includegraphics[width=\linewidth]{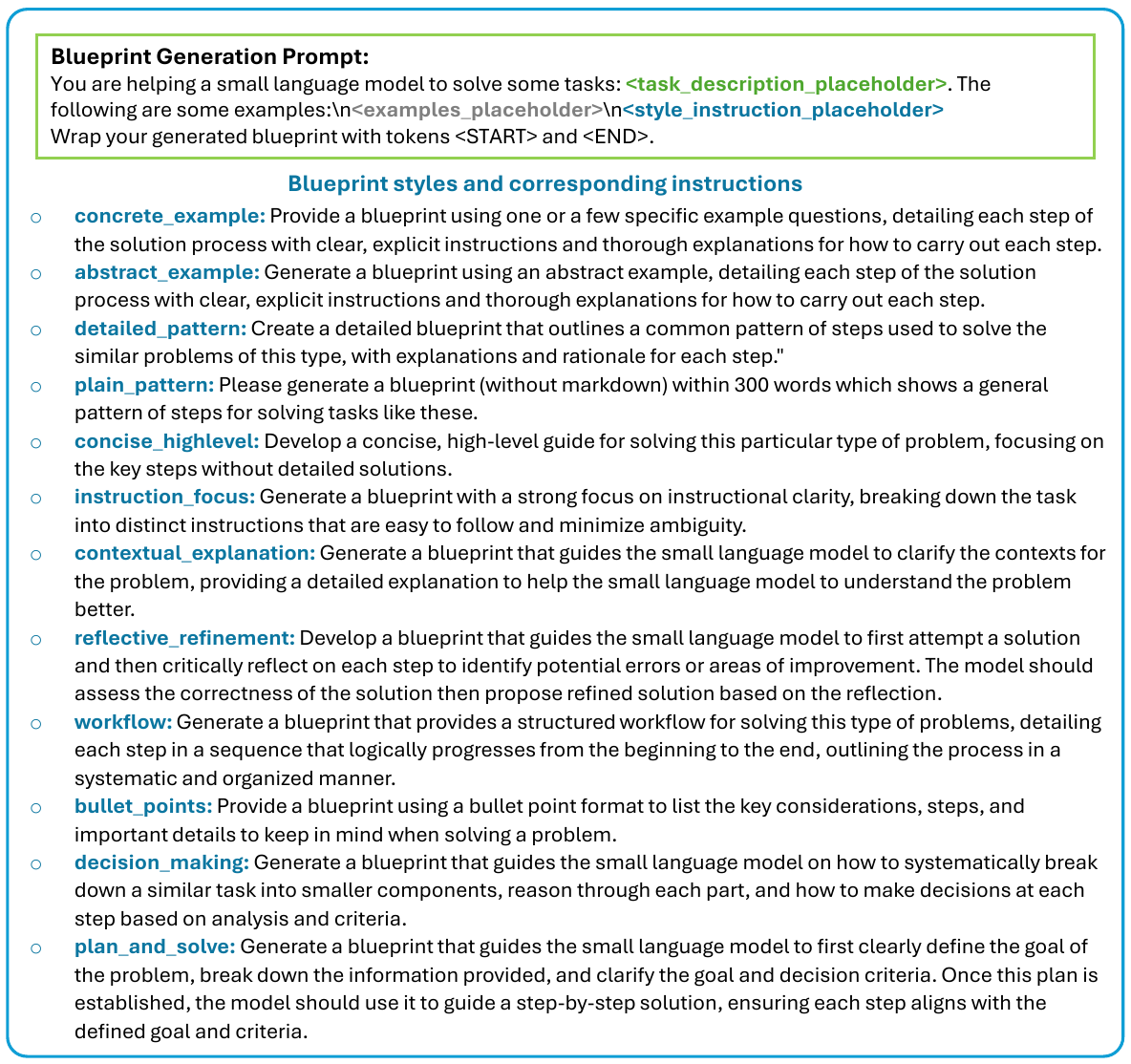}
    \caption{Blueprint generation prompt, blueprint styles and their corresponding generating instructions. When generating a blueprint, take a blueprint style and the corresponding generating instruction, which replaces the \texttt{<style-instruction-placeholder>} to form part of the blueprint generation prompt. The prompt is then used as an input to the LLM for generating the blueprint.}
    \label{fig:blueprint_styles}
\end{figure*}

Fig.~\ref{fig:blueprint_styles} shows the blueprint generation prompt, accompanied by the 12 blueprint generation styles and their corresponding instruction prompts. The styles are chosen to reflect a diverse selection of different reasoning styles and formats. For example, the blueprints generated with styles \concreteexample{} and \abstractexample{} will use concrete examples or synthesized examples to illustrate detailed reasoning steps for solving a problem according to the general pattern of problem-solving for the task category. The styles \detailed{}, \plain{}, and \concise{} present the SLM with the instruction of pattern of steps of reasoning, in decreasing level of details, moreover, \bulletpoint{} will present the guidance in detailed steps and in bullet-pointed format. \instruction{} and \contextual{} will focus on clarity by providing clarified and unambiguous instructions for the SLM to understand the problem context and reasoning steps. The \reflective{}, \decision{} and \plan{} blueprints aim to guide an SLM to perform multi-step reasoning in a single turn, such as reflection and refinement, or planning steps before formulating the solution.

In our experiments, the blueprint style selection is performed with $10$ training examples for each of the $12$ blueprint styles. Therefore, the number of SLM calls performed during blueprint style selection is $10 \times 12 
 = 120$ SLM calls.

\subsection{More details on Blueprint APO Refinement}
\label{sec:appendix_blueprint_apo}
In this section, we include more details on the blueprint APO refinement procedure, as well as the detailed prompts used by $\llm{}$ for generating and refining the blueprint.

For a given task category $\task$ and SLM $\slm$, each APO round starts with an initial blueprint $\bp$ (or a set of candidate blueprints) and works as follows:
\begin{enumerate}[noitemsep,topsep=0pt]
\item \textbf{Initial blueprint evaluation:} Evaluate $\slm$ using the initial blueprint on sampled training data (in our experiments, we used 25 training examples).
\item \textbf{Compile error message:} Examples that $\slm$ answered incorrectly are compiled as an error message $e$ with the questions, $\slm$ responses and correct solutions.
\item \textbf{Textual gradient:} Prompt $\llm$ with the task description, blueprint $\bp$, and error message $e$ to generate a list of \emph{textual gradients} $\{g_j\}_{j=0}^{N_{grad}}$, i.e., each textual gradient is an error analysis of how the blueprint may have caused the error examples. 
\item \textbf{Edit:} For each gradient $g_j$, prompt $\llm$ with the initial blueprint $\bp$, the error message $e$ and gradient $g_j$ to generate a refined blueprint $\bp^{edit}_j$.
\item \textbf{Paraphrase:} Prompt $\llm$ to write a paraphrased blueprint $\bp^{para}_j$ for each refined blueprint $\bp^{edit}_j$.
\item \textbf{Select:} Evaluate $\slm$ with the initial blueprint and all refined and paraphrased blueprints on sampled training data, selecting the best performing candidate(s). 
\end{enumerate}
The APO round can be repeated for multiple iterations with beam search, where the expansion steps (Step 1 to 5) expand from the initial blueprint candidates, followed by the selection step (Step 6) narrowing down to $N_{beams}$ blueprint candidates for the next round.

In our experiments, we perform 1 round of APO, starting with the blueprint with the best style from the blueprint generation step. For Step 1 (initial blueprint evaluation), we perform evaluation of the $\slm{}$ on 25 training examples, and sample a maximum of 5 error examples for compiling the error message in Step 2. For Step 3, we generate 2 textual gradients, thereby obtaining 2 refined blueprints in Step 4, and 2 new paraphrased blueprints. In Step 6 of selection, we perform selection of the one top scoring blueprint. Due to the relatively small candidate size, the selection is performed with 20 training examples per blueprint candidate. The total number of SLM calls in the APO process is then $25 + 20 \times 5 = 125$ SLM calls. And the total number of LLM calls in the APO process is $6$, with $2$ calls per Step 3-5.

\begin{figure*}[t]
    \centering
    \includegraphics[width=\linewidth]{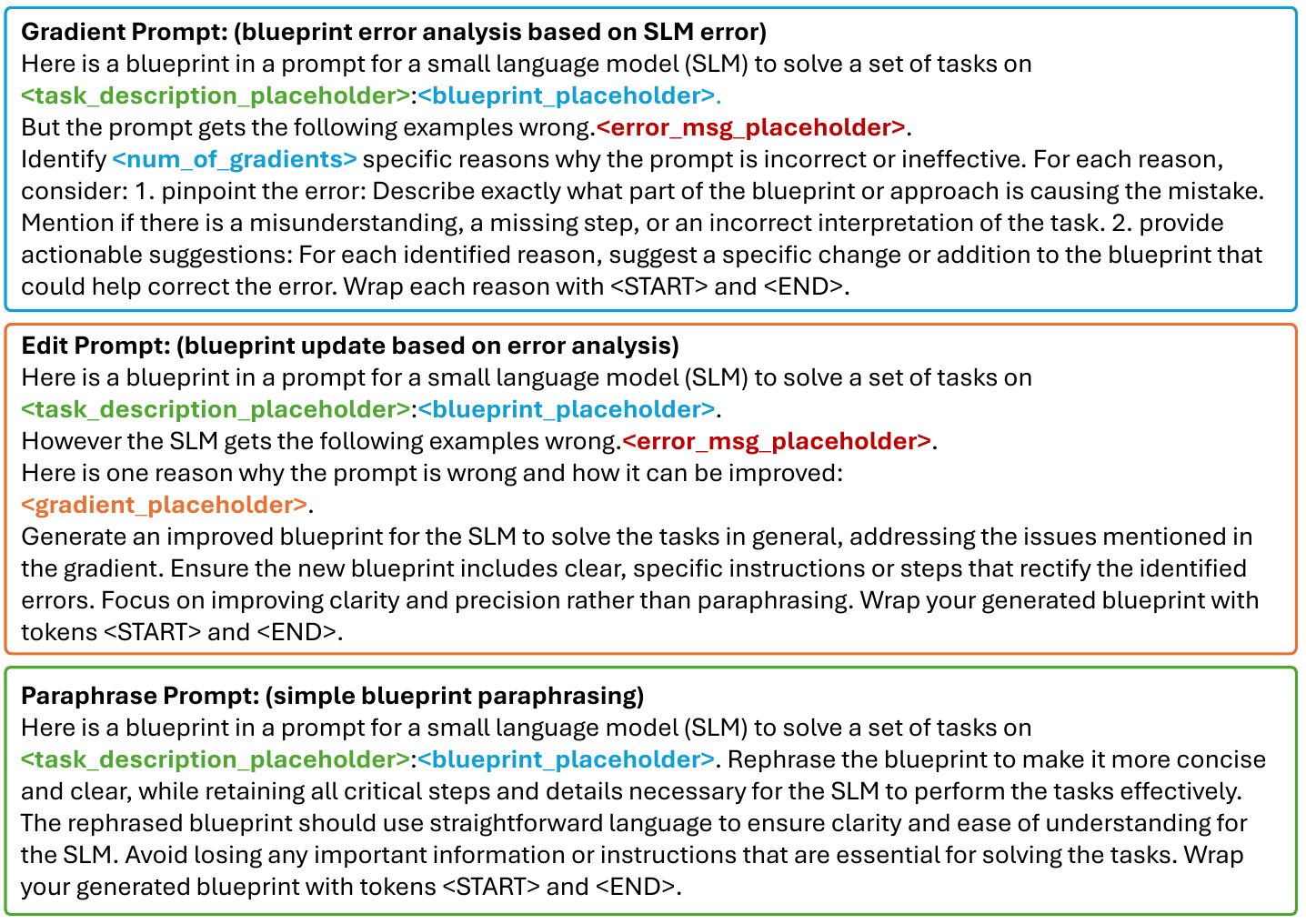}
    \caption{Prompts used for blueprint refinement via Automatic Prompt Optimization (APO). During each APO round, the LLM first performs error analysis and generates a \emph{textual gradient} using the Gradient Prompt, then refines and updates the blueprint according to the textual gradient with the Edit Prompt. Finally, the LLM paraphrase the blueprint with the Paraphrase Prompt. During execution, the placeholders are replaced with their actual values.}
    \label{fig:blueprint_generation_apo_prompts}
\end{figure*}

This optimization procedure can help us to obtain a blueprint that is refined according to the behaviors of $\slm$ on task $\task$.
All prompts used for generating the textual gradient, editing and updating the blueprint, as well as paraphrasing the blueprint are included in Fig.~\ref{fig:blueprint_generation_apo_prompts}.
For more details on APO please refer to~\cite{pryzant2023automatic}.

\section{More Details on Template Search}
\label{sec:appendix_template}
To perform prompt template search, we define the search space as shown in Fig.~\ref{fig:framework}, with 4 template parameters:
\begin{itemize}[noitemsep,topsep=0pt]
\item Number of In-context examples = [0,1,2,3]
\item Placing Task Description before In-context Examples = [True, False]
\item Include Blueprint = [True, False]
\item Include CoT = [True, False]
\end{itemize}
Specifically, the in-context examples are provided by the respective datasets, which includes step-by-step solutions. The inclusion of CoT (chain-of-thought) means the inclusion of the sentence \emph{Let's think step-by-step}. In total the prompt template parameters produce $32$ combinations of different prompt templates. For example, if we choose 3 in-context examples, place task description before in-context examples, include the blueprint and not include CoT, then the prompt template produced is the following: \texttt{<task-description><3-in-context-examples>} \texttt{<blueprint><question>}. 
In our experiments, we perform the successive halving template search with a reduction factor of $f=2$, (i.e., we evaluate $32$ templates in the initial iteration, then successively reduce by half in subsequent iterations until we obtain the top performing candidate). At each iteration, we sample $5$ training examples for evaluating each of remaining the prompt template candidates, by applying each prompt template to form the SLM prompt. In total, the number of SLM calls incurred during prompt template search per task category per SLM model is $310$ SLM calls.

\section{Detailed Per-category Results on BBH}
\label{sec:appendix_bbh}
\begin{figure*}[t]
\vspace{-4mm}
    \captionsetup[subfigure]{skip=-5pt}
    \begin{subfigure}[b]{0.24\textwidth}
        \includegraphics[width=\textwidth, height=0.8\textwidth]{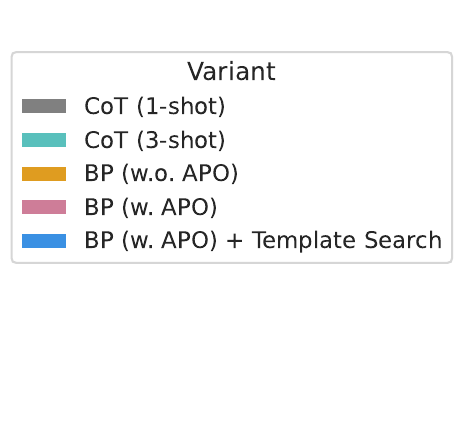}
    \end{subfigure}
    \begin{subfigure}[b]{0.24\textwidth}
        \includegraphics[width=\textwidth, height=0.8\textwidth]{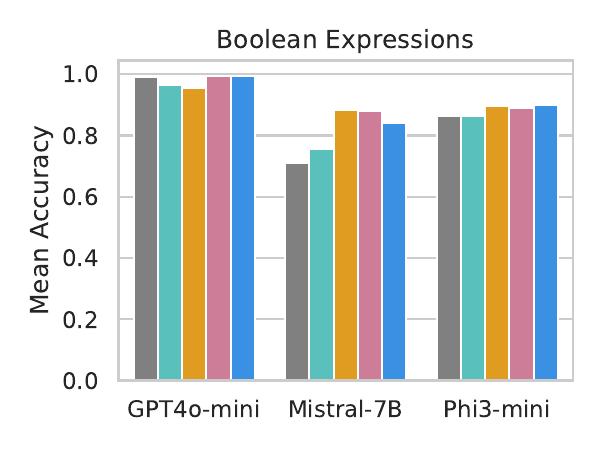}
        \caption{Boolean Expressions}
        \label{fig:bbh_boolean_expressions}
    \end{subfigure}
   \begin{subfigure}[b]{0.24\textwidth}
        \includegraphics[width=\textwidth, height=0.8\textwidth]{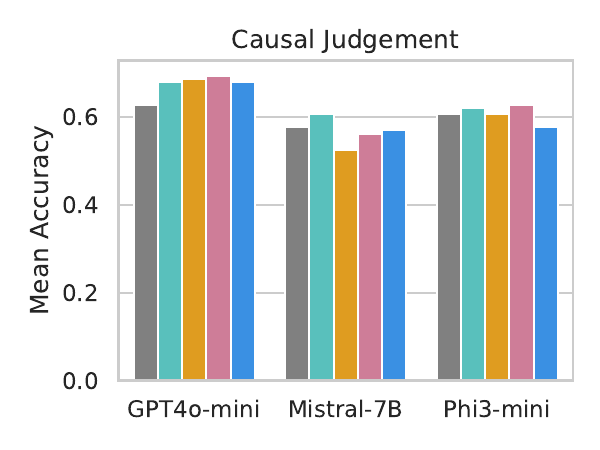}
        \caption{Causal Judgement}
        \label{fig:bbh_causal_judgement}
    \end{subfigure}
    \begin{subfigure}[b]{0.24\textwidth}
        \includegraphics[width=\textwidth, height=0.8\textwidth]{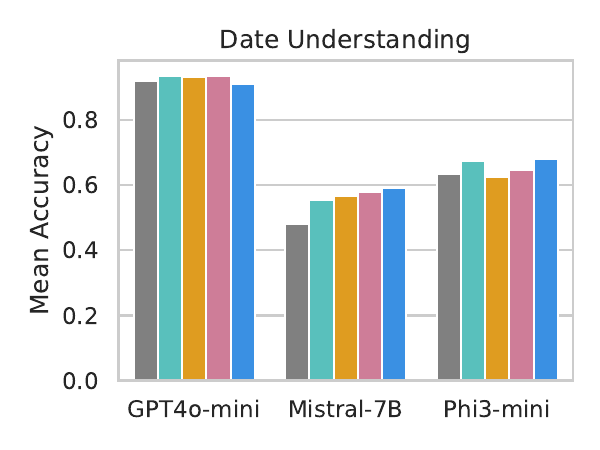}
        \caption{Date Understanding}
        \label{fig:bbh_date_understanding}
    \end{subfigure}
    \begin{subfigure}[b]{0.24\textwidth}
        \includegraphics[width=\textwidth, height=0.8\textwidth]{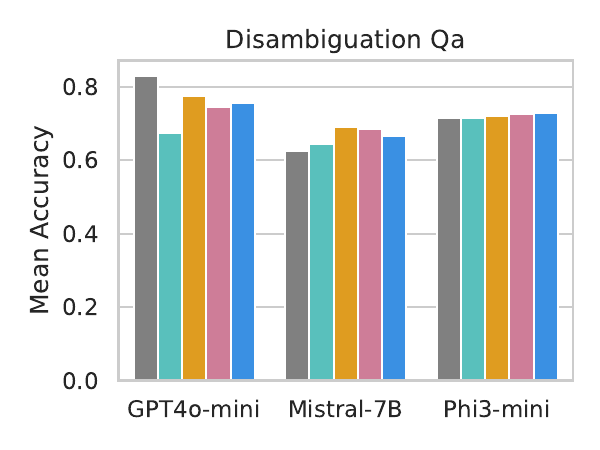}
        \caption{Disambiguation QA}
        \label{fig:bbh_disambiguation_qa}
    \end{subfigure}
    \begin{subfigure}[b]{0.24\textwidth}
        \includegraphics[width=\textwidth, height=0.8\textwidth]{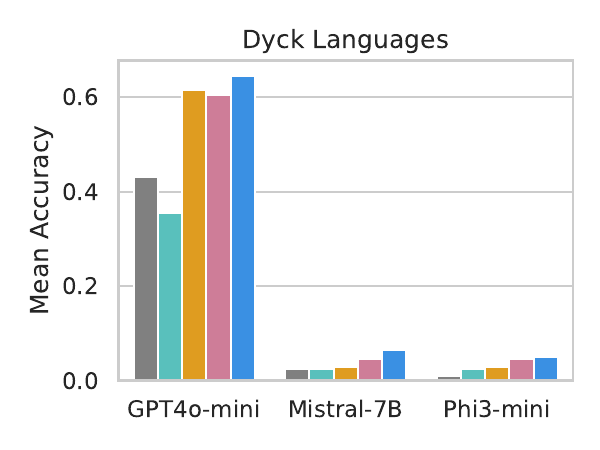}
        \caption{Dyck Languages}
        \label{fig:bbh_dyck_languages}
    \end{subfigure}
    \begin{subfigure}[b]{0.24\textwidth}
        \includegraphics[width=\textwidth, height=0.8\textwidth]{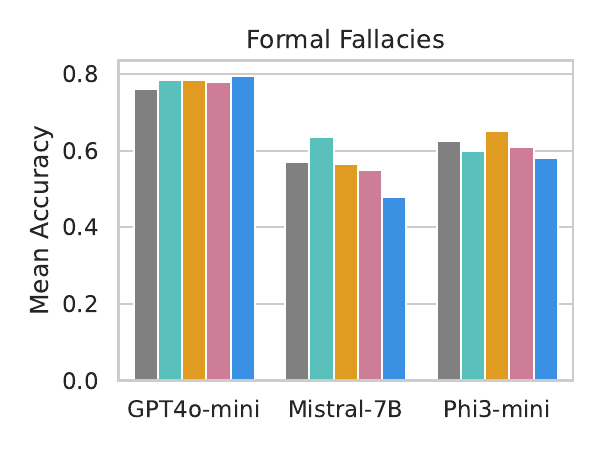}
        \caption{Formal Fallacies}
        \label{fig:bbh_formal_fallacies}
    \end{subfigure}
    \begin{subfigure}[b]{0.24\textwidth}
        \includegraphics[width=\textwidth, height=0.8\textwidth]{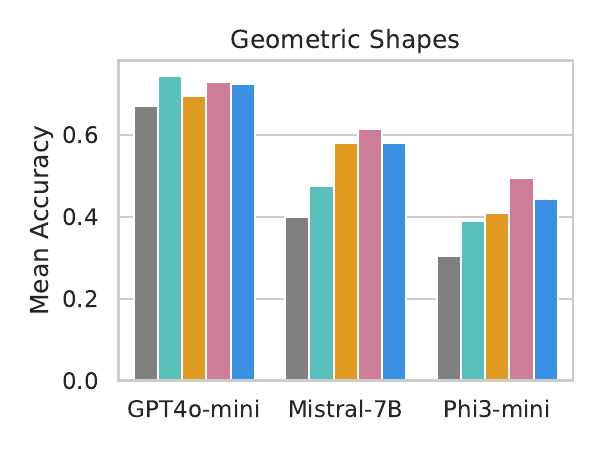}
        \caption{Geometric Shapes}
        \label{fig:bbh_geometric_shapes}
    \end{subfigure}
    \begin{subfigure}[b]{0.24\textwidth}
        \includegraphics[width=\textwidth, height=0.8\textwidth]{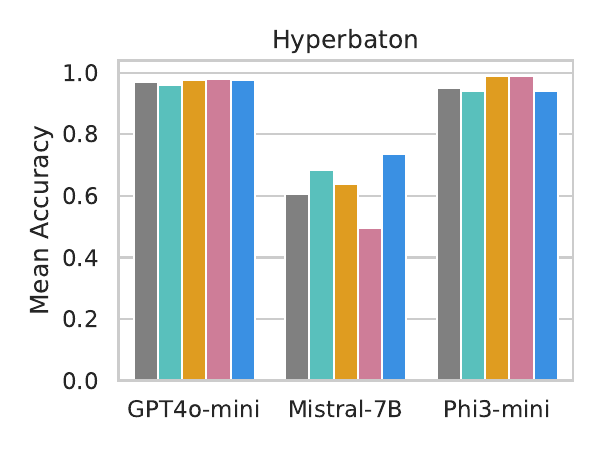}
        \caption{Hyperbaton}
        \label{fig:bbh_hyperbaton}
    \end{subfigure}
    \begin{subfigure}[b]{0.24\textwidth}
        \includegraphics[width=\textwidth, height=0.8\textwidth]{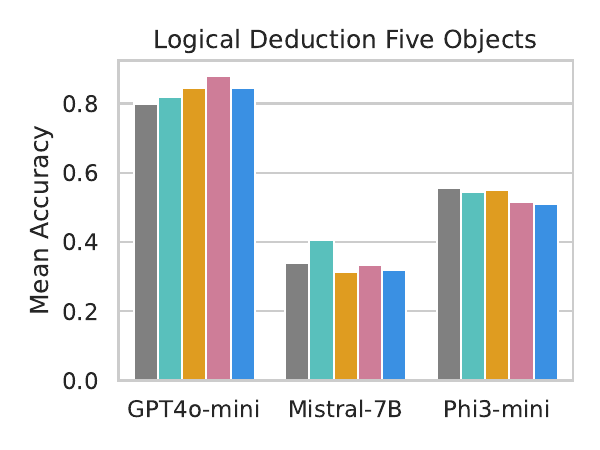}
        \caption{Logical Deduction (five)}
        \label{fig:bbh_logical_deduction_five_objects}
    \end{subfigure}
    \begin{subfigure}[b]{0.24\textwidth}
        \includegraphics[width=\textwidth, height=0.8\textwidth]{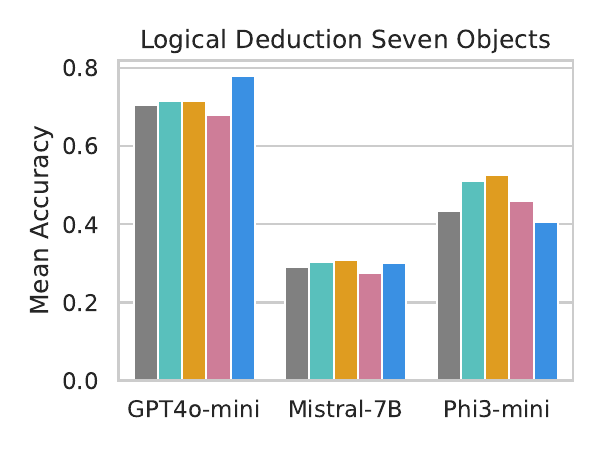}
        \caption{Logical Deduction (seven)}
        \label{fig:bbh_logical_deduction_seven_objects}
    \end{subfigure}
    \begin{subfigure}[b]{0.24\textwidth}
        \includegraphics[width=\textwidth, height=0.8\textwidth]{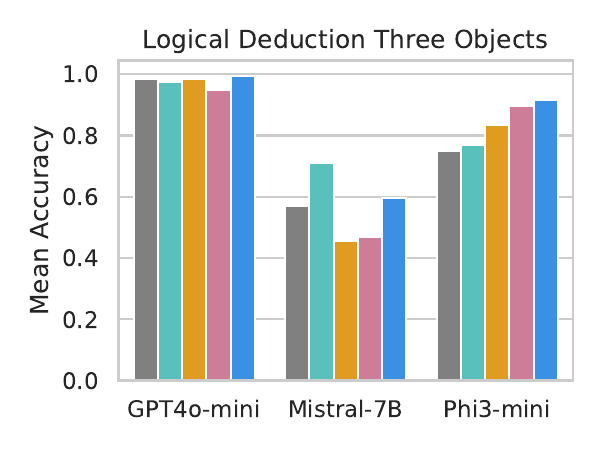}
        \caption{Logical Deduction (three)}
        \label{fig:bbh_logical_deduction_three_objects}
    \end{subfigure}
    \begin{subfigure}[b]{0.24\textwidth}
        \includegraphics[width=\textwidth, height=0.8\textwidth]{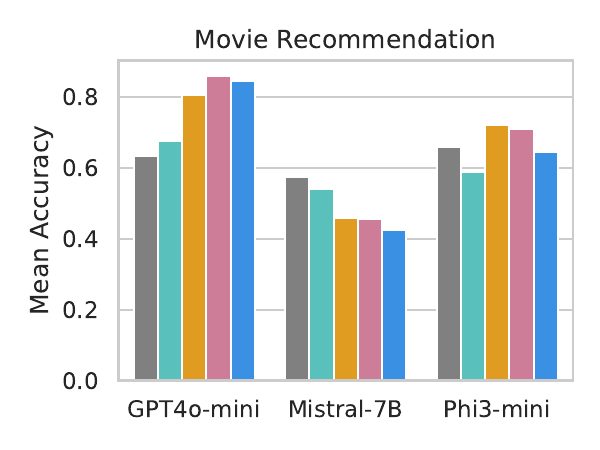}
        \caption{Movie Recommendation}
        \label{fig:bbh_movie_recommendation}
    \end{subfigure}
    \begin{subfigure}[b]{0.24\textwidth}
        \includegraphics[width=\textwidth, height=0.8\textwidth]{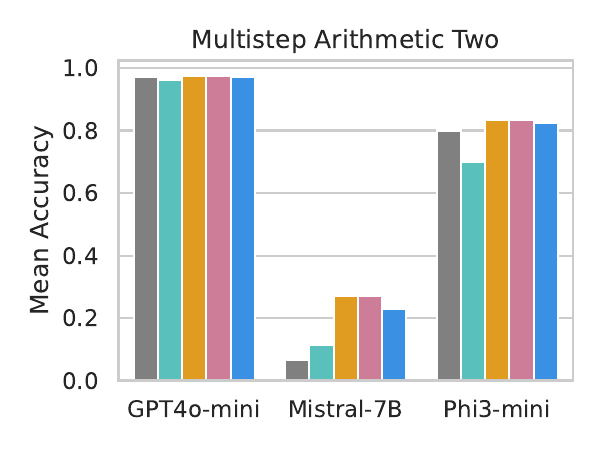}
        \caption{Multistep Arithmetic Two}
        \label{fig:multistep_arithmetic_two}
    \end{subfigure}
    \begin{subfigure}[b]{0.24\textwidth}
        \includegraphics[width=\textwidth, height=0.8\textwidth]{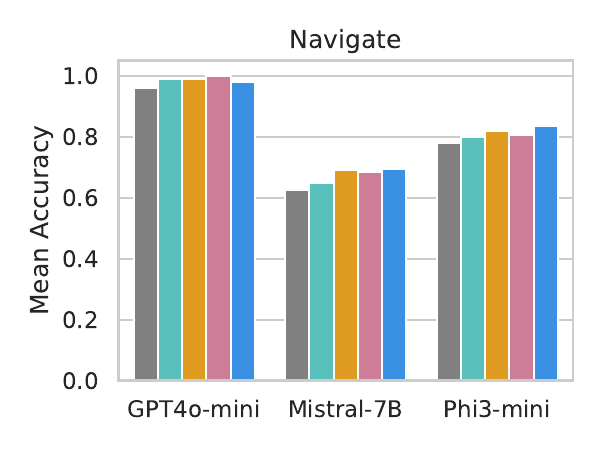}
        \caption{Navigate}
        \label{fig:navigate}
    \end{subfigure}
    \begin{subfigure}[b]{0.24\textwidth}
        \includegraphics[width=\textwidth, height=0.8\textwidth]{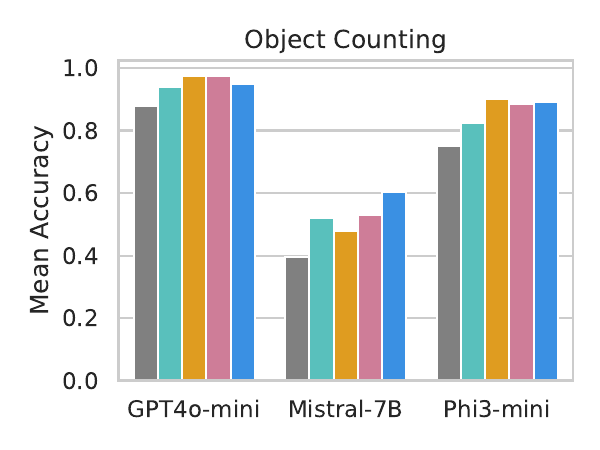}
        \caption{Object Counting}
        \label{fig:object_counting}
    \end{subfigure}
    \begin{subfigure}[b]{0.24\textwidth}
        \includegraphics[width=\textwidth, height=0.8\textwidth]{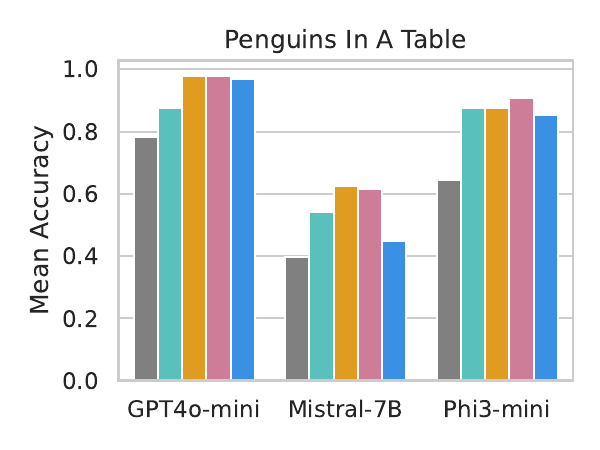}
        \caption{Penguins In A Table}
        \label{fig:penguins_in_a_table}
    \end{subfigure}
    \begin{subfigure}[b]{0.24\textwidth}
        \includegraphics[width=\textwidth, height=0.8\textwidth]{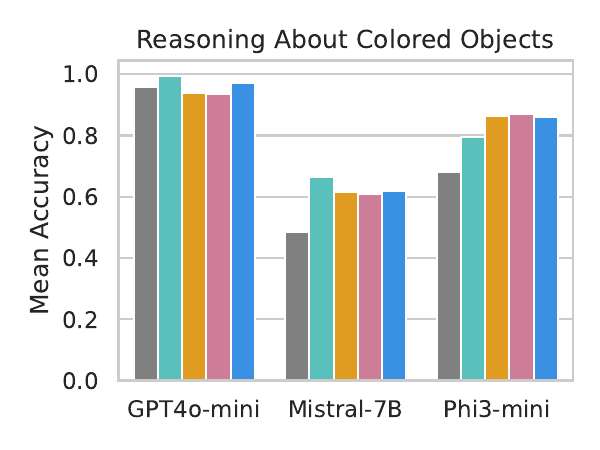}
        \caption{Reasoning Colored Objects}
        \label{fig:reasoning_about_colored_objects}
    \end{subfigure}
    \begin{subfigure}[b]{0.24\textwidth}
        \includegraphics[width=\textwidth, height=0.8\textwidth]{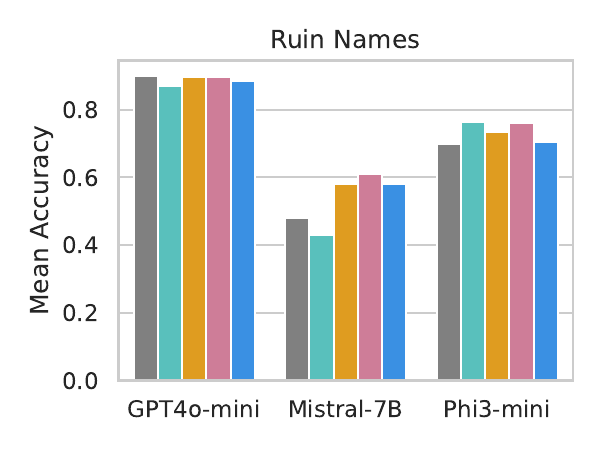}
        \caption{Ruin Names}
        \label{fig:ruin_names}
    \end{subfigure}
    \begin{subfigure}[b]{0.24\textwidth}
        \includegraphics[width=\textwidth, height=0.8\textwidth]{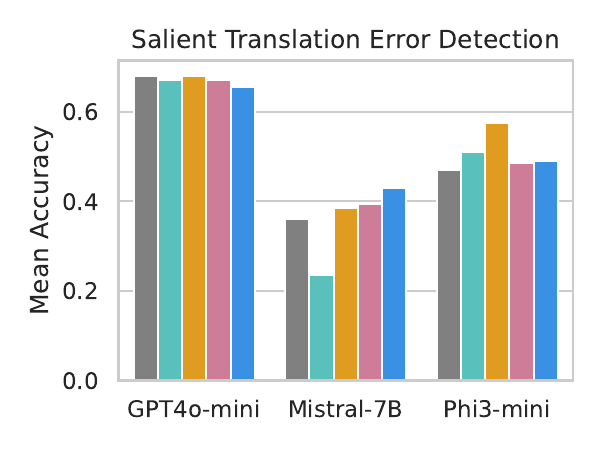}
        \caption{Salient Translation}
        \label{fig:salient_translation_error_detection}
    \end{subfigure}
    \begin{subfigure}[b]{0.24\textwidth}
        \includegraphics[width=\textwidth, height=0.8\textwidth]{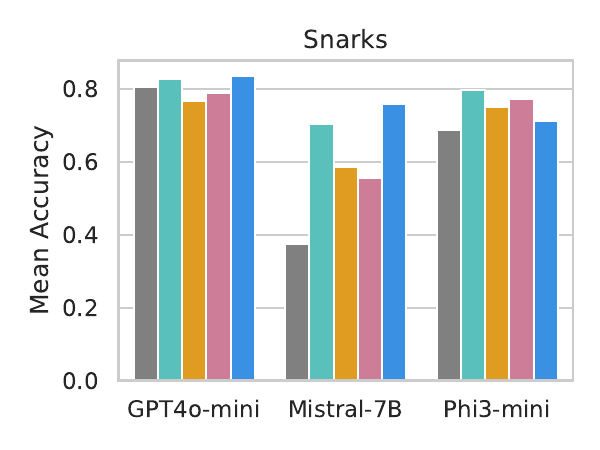}
        \caption{Snarks}
        \label{fig:snarks}
    \end{subfigure}
     \begin{subfigure}[b]{0.24\textwidth}
        \includegraphics[width=\textwidth, height=0.8\textwidth]{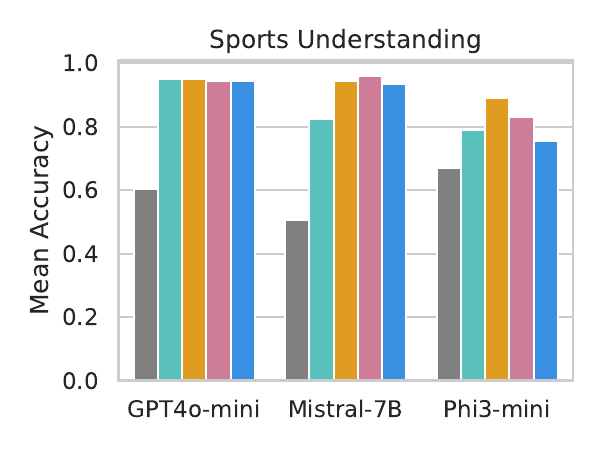}
        \caption{Sports Understanding}
        \label{fig:sports_understanding}
    \end{subfigure}
    \begin{subfigure}[b]{0.24\textwidth}
        \includegraphics[width=\textwidth, height=0.8\textwidth]{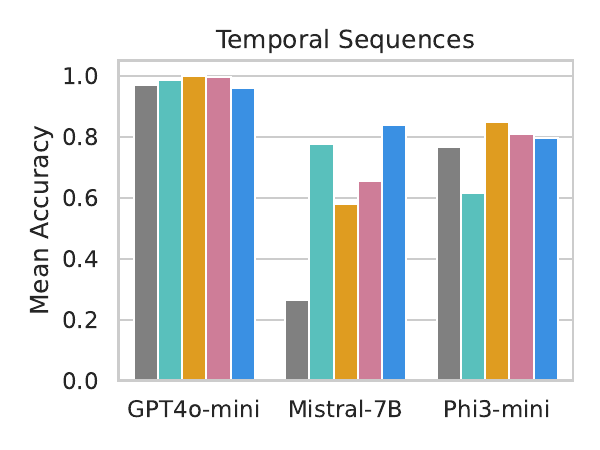}
        \caption{Temporal Sequences}
        \label{fig:temporal_sequences}
    \end{subfigure}
    \begin{subfigure}[b]{0.24\textwidth}
        \includegraphics[width=\textwidth, height=0.8\textwidth]{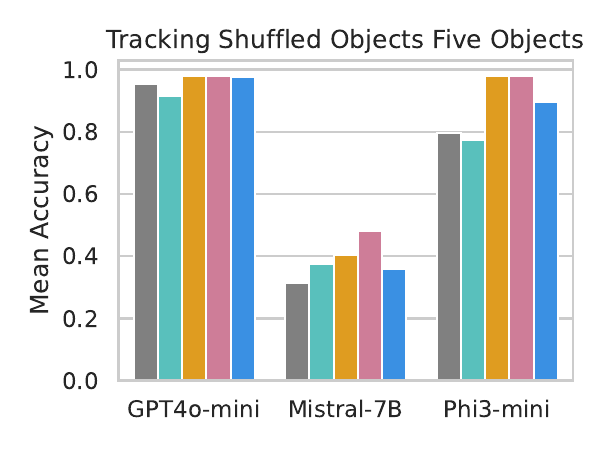}
        \caption{Tracking Shuffled Five}
        \label{fig:tracking_shuffled_objects_five_objects}
    \end{subfigure}
    \begin{subfigure}[b]{0.24\textwidth}
        \includegraphics[width=\textwidth, height=0.8\textwidth]{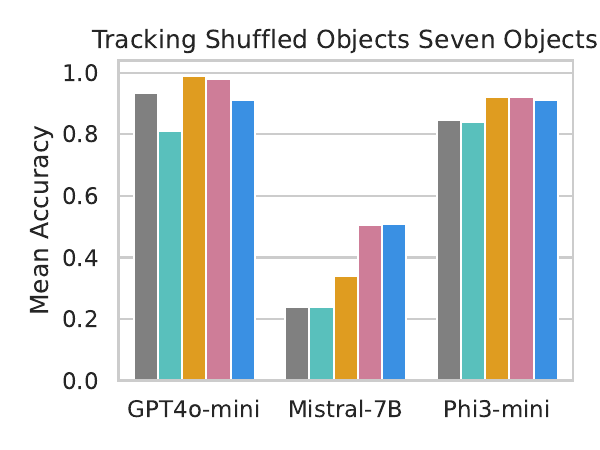}
        \caption{Tracking Shuffled Seven}
        \label{fig:tracking_shuffled_objects_seven_objects}
    \end{subfigure}
    \begin{subfigure}[b]{0.24\textwidth}
        \includegraphics[width=\textwidth, height=0.8\textwidth]{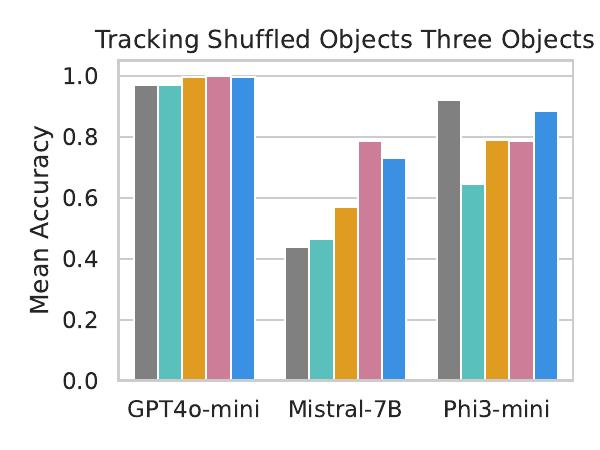}
        \caption{Tracking Shuffled Three}
        \label{fig:tracking_shuffled_objects_three_objects}
    \end{subfigure}
    \begin{subfigure}[b]{0.24\textwidth}
        \includegraphics[width=\textwidth, height=0.8\textwidth]{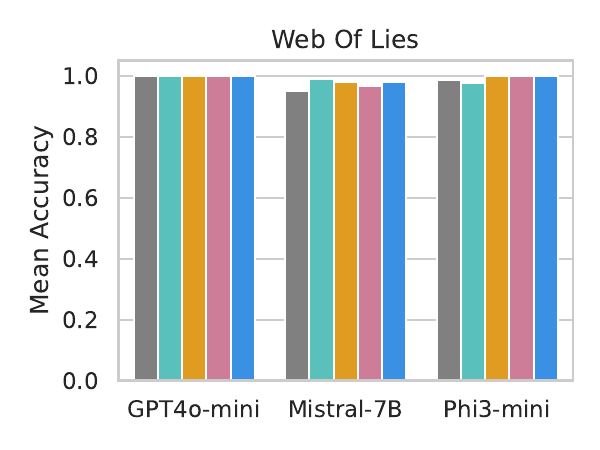}
        \caption{Web of Lies}
        \label{fig:web_of_lies}
    \end{subfigure}
    \caption{Per-category Performance on the BBH dataset}
    \label{fig:bbh_all}
\end{figure*}

Fig.~\ref{fig:bbh_all} shows a detailed comparison of the different method variants on each sub-category of the BBH dataset. Each plot corresponds to a task category, which shows the performance of the three SLM models (\gptmini{}, \mistral{}, \phimini{}) using different method variants. 

Overall we can observe a trend where the mean accuracy increase from left to right (the leftmost two bars corresponds to the \cotoneshot{} and \cotthreeshot{} baselines, while the other three bars correspond to our methods, BP w.o. APO, BP w. APO and BP w. APO + Template Search, respectively).
Comparing our method with \cotthreeshot{}, we can observe significant task improvement in several categories. For example, on Dyck Languages (Fig.~\ref{fig:bbh_dyck_languages}), \gptmini{} shows $25\%$ performance improvement with \bpwapo{} and a further $4\%$ improvement with template search. On Geometric Shapes (Fig.~\ref{fig:bbh_geometric_shapes}), \mistral{} shows a $14\%$ with \bpwapo{} and $10.5\%$ improvement with \bpwapotemplatesearch{}. On Logical Deduction Three Objects (Fig.~\ref{fig:bbh_logical_deduction_three_objects}), \phimini{} shows a $12.5\%$ improvement with \bpwapo{} and a further $2\%$ improvement with template search. Sometimes template search may not find the optimal template with the highest average accuracy across all variants, as a result of limited number of training data for evaluating the templates. Nevertheless, the template search procedure can usually identify a template with near-optimal performance as well as preventing low performing templates. Therefore we observe that it displays the most robust and overall best performance across all task categories and SLMs.

\section{More Probing Experiment Results on Prompt Sensitivity}
\label{sec:appendix_probing}
\begin{figure*}[t]
    \captionsetup[subfigure]{skip=-7pt}
    \begin{subfigure}[b]{0.22\textwidth}
        \includegraphics[width=\textwidth, height=0.9\textwidth]{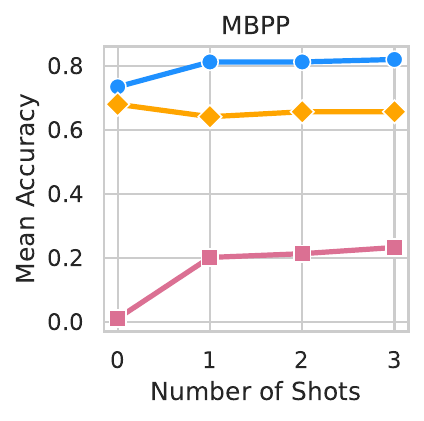}
        \label{fig:shots_mbpp}
    \end{subfigure}
    \hspace{-2mm}
    \begin{subfigure}[b]{0.22\textwidth}
        \includegraphics[width=\textwidth, height=0.9\textwidth]{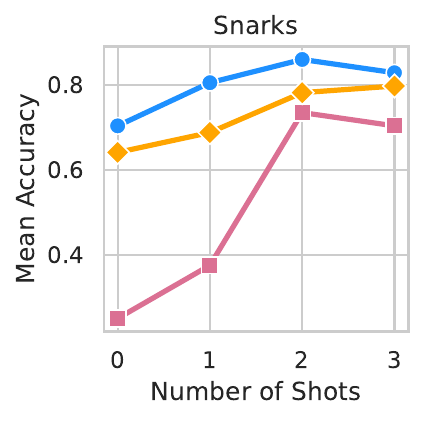}
        \label{fig:shots_snarks}
    \end{subfigure}
    \hspace{-2mm}
    \begin{subfigure}[b]{0.22\textwidth}
        \includegraphics[width=\textwidth, height=0.9\textwidth]{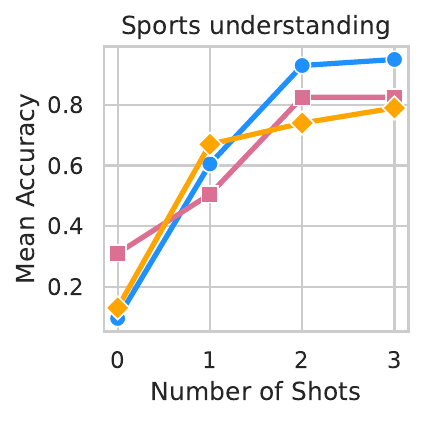}
        \label{fig:shots_sports}
    \end{subfigure}
    \hspace{-2mm}
    \begin{subfigure}[b]{0.22\textwidth}
        \includegraphics[width=\textwidth, height=0.9\textwidth]{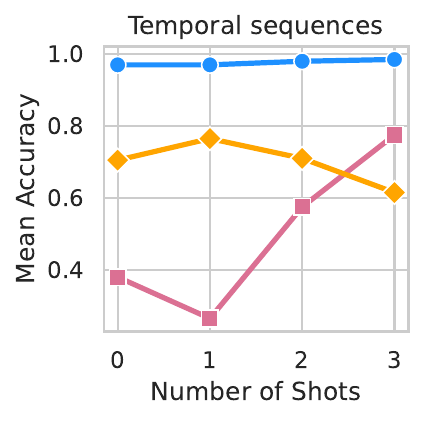}
        \label{fig:shots_temporal}
    \end{subfigure}
    \hspace{-2mm}
    \begin{subfigure}[b]{0.12\textwidth}
        \includegraphics[width=\textwidth, height=1.5\textwidth]{latex/img/exp/sensitivity/legend.pdf}
        \label{fig:shots_legend}
    \end{subfigure}
    \caption{Sensitivity of SLMs to prompt variations across tasks - the number of in-context examples can significantly impact the SLM performance across different tasks. For instance in the temporal sequences dataset (BBH), \gptmini{} shows relatively stable a d slightly improved performance with increasing number of examples, \mistral performance dropped from 0-shot to 1-shot, while the performance improved from 1-shot to 3-shot, whereas \phimini performed best with 1-shot example and its performance dropped from 1-shot to 3-shot examples. In contrast, on Sports understanding dataset (BBH), all models show improved performance with increasing number of in-context examples.
    }
    \label{fig:shots}
\end{figure*}

Apart from observing that the changing the ordering of task description and in-context examples can significantly change task performance, we also present the effect of varying the number of in-context examples on the task performance on a subset of BBH datasets. In Figure~\ref{fig:shots}, we observe that increasing the number of in-context examples from 0 to 3 often results in improved performance, for example, on MBPP both \gptmini{} and \mistral{} shows improved performance while \phimini{}'s performance is relatively stable. For Sports Understanding (BBH), all SLMs show significantly improved performance with increasing number of examples. Sometimes increasing number of examples may results in lower performance, for example, in Temporal Sequence (BBH), \phimini{} shows increasing performance as we increase the number of examples from 0 to 1, while we see decreasing performance as we increase the number of examples from 1 to 3, whereas \mistral{} displays opposite trends.

\section{More Details on Experiment Setup}
\label{sec:appendix_experiment_details}
The sizes of the SLMs are: \phimini{} has 3.8B parameters, \mistral{} has 7B parameters, we do not know the size of the \gptmini{} model. For each SLM model, we use the same hyper-parameter of temperature=0 and top-p=1. For a fair comparison, we use the same hyper-parameters across all the APO variants (including our variants that optimize the blueprint and the baseline variant that optimizes the task description), details on the APO steps, hyper-parameters, and number of LLM/SLM calls are presented in Section~\ref{sec:appendix_blueprint_apo}. Details on template search are presented in Section~\ref{sec:appendix_template}. For each dataset within GSM8K, MBPP, and each sub-category of BBH, we use 50 training examples, and evaluate all variants on the same held-out test set. For efficient evaluation on a wide task diversity, we use 200-300 evaluation datapoints across a total of 28 task categories and report the average accuracy across all datapoints for each category (except for BBH which we report the average over all sub-categories in Table~\ref{table: main}), therefore, the total datapoints evaluated for each SLM and method variant is around 5600). In detail, for GSM8K, we used the same 300 evaluation datapoints from Huggingface for evaluating all SLM/method combinations. For MBPP, we used the sanitized release of MBPP dataset from Huggingface, including 257 evaluation datapoints. For BBH, we use the dataset from Huggingface, with 50 training datapoints, and 200 evaluation datapoints per sub-category (for two sub-categories with fewer than 200 evaluation datapoints, we use all available datapoints). We used regular expressions for efficient evaluation on multiple-choice and calculation problems (26 out of 27 BBH sub-categories), where the expected answers are well-defined. The word sorting category was excluded due to the variability in natural language responses that could not be evaluated reliably using regular expressions.

\end{document}